\def\year{2020}\relax
\documentclass[letterpaper]{article} 
\usepackage{aaai20}  
\usepackage{times}  
\usepackage{helvet} 
\usepackage{courier}  
\usepackage[hyphens]{url}  
\usepackage{graphicx} 
\usepackage{graphicx}  
\usepackage{amsmath}
\usepackage{amssymb}
\usepackage{booktabs}
\usepackage{multirow}
\usepackage{array}

\title{Distraction-Aware Feature Learning for Human Attribute Recognition via Coarse-to-Fine Attention Mechanism}

\author{Mingda Wu,\textsuperscript{\rm 1} Di Huang,\textsuperscript{\rm 1}\thanks{Corresponding author: Di Huang.} Yuanfang Guo,\textsuperscript{\rm 2} Yunhong Wang\textsuperscript{\rm 1}\\ 
\textsuperscript{\rm 1}Beijing Advanced Innovation Center for Big Data and Brain Computing, Beihang University, Beijing, China\\
\textsuperscript{\rm 2}IRIP Lab,  School of Computer Science and Engineering, Beihang University, Beijing, China\\
\{md99504, dhuang, andyguo, yhwang\}@buaa.edu.cn 
}
 
\begin{document}

\maketitle

\begin{abstract}
Recently, Human Attribute Recognition (HAR) has become a hot topic due to its scientific challenges 
and application potentials, where localizing attributes is a crucial stage but not well handled. In this 
paper, we propose a novel deep learning approach to HAR, namely Distraction-aware HAR (Da-HAR). 
It enhances deep CNN feature learning by improving attribute localization through a coarse-to-fine 
attention mechanism. At the coarse step, a self-mask block is built to roughly discriminate and reduce 
distractions, while at the fine step, a masked attention branch is applied to further eliminate irrelevant 
regions. Thanks to this mechanism, feature learning is more accurate, especially when heavy 
occlusions and complex backgrounds exist. Extensive experiments are conducted on the 
WIDER-Attribute and RAP databases, and state-of-the-art results are achieved, demonstrating 
the effectiveness of the proposed approach.
\end{abstract}

\section{Introduction}

Given an input image with a target person, Human Attribute Recognition (HAR) predicts his or her 
semantic characteristics, including low-level ones (\emph{e.g.} wearing logo or plaid), mid-level 
ones (\emph{e.g.} wearing hat or T-shirt), and high-level ones (\emph{e.g.} gender, dressing formally). 
Accurate recognition of such attributes not only improves machine intelligence on cognition of humans, 
but also benefits a large number of applications such as person re-identification 
\cite{ling2019improving,han2018attribute}, pedestrian detection \cite{tian2015pedestrian}, and person 
retrieval \cite{feris2014attribute,wang2013personal}.

\begin{figure}
\begin{center}
   \includegraphics[width=1.0\linewidth]{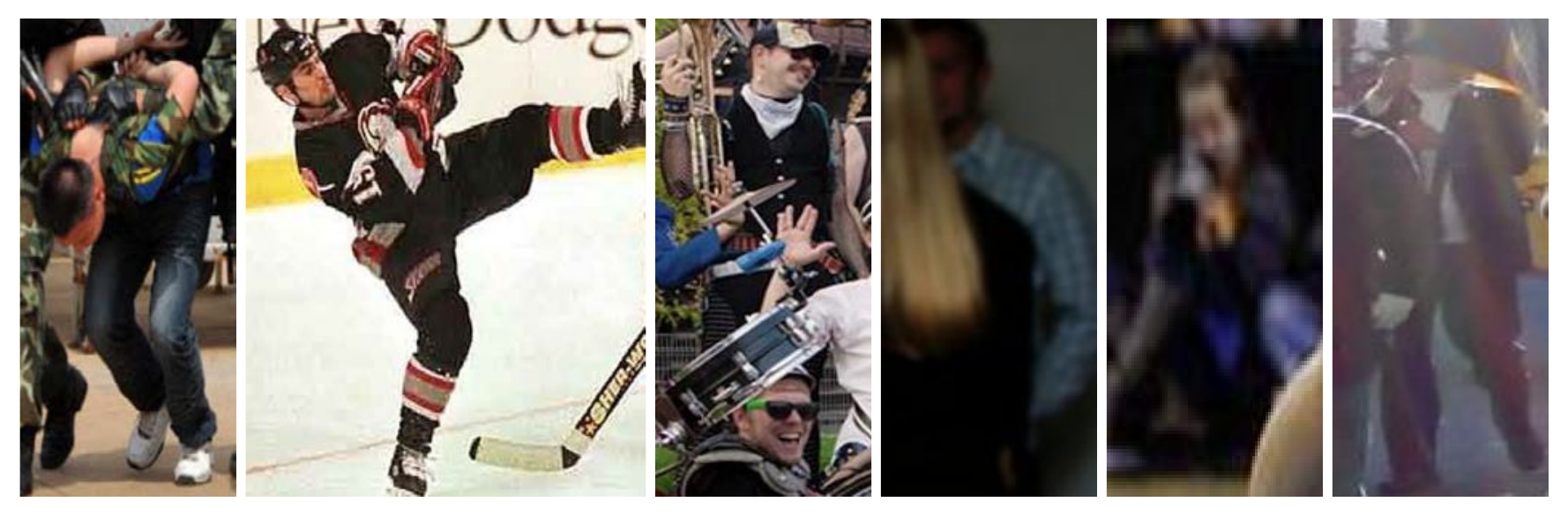}
\end{center}
   \caption{Human Attribute Recognition is challenging due to large variations of body gestures, 
   external occlusions, lighting conditions, image resolutions and blurrinesses.}
\label{fig:fig1}
\end{figure}

Existing investigations of HAR can be classified into three domains, \emph{i.e.}, clothing domain, 
surveillance domain, and general domain. The techniques in the clothing domain have received 
extensive attentions \cite{al2017fashion,sarafianos2017adaptive,liu2016deepfashion,chen2015deep} 
due to their potentials in commercial applications. This type of methods generally require the input 
images of high resolutions with persons at a small number of pre-defined poses, and fine-grained 
clothing style recognition is still challenging. There are also numerous studies in the surveillance 
domain \cite{DCL,gao2019pedestrian,VSGR,LGNet,JRL}, because these techniques are playing an 
important role in public security. Input images are recorded by a diversity of monitoring cameras, 
and the major difficulties lie in low resolutions, high blurrinesses, and complex backgrounds. 
In the past several years, interests have been shown in the general domain \cite{CAM,VeSPA,WIDER}, 
where input images are acquired in arbitrary scenarios exhibiting additional variations of gestures, 
viewpoints, illuminations, and occlusions, as depicted in Figure \ref{fig:fig1}.

Regardless of differences between domains, HAR methods basically share a common framework, 
which conducts attribute-sensitive feature extraction on attribute-related regions for classification. 
In the literature, the majority of existing efforts to HAR have been made on building effective features, 
and a large number of works focus on improving the discrimination and the robustness of 
representations of appearance properties. Features are evolving from handcrafted ones 
\cite{joo2013human,cao2008gender} to deep learned ones \cite{zhu2017multi,WPAL}, with promising 
performance achieved. 

\begin{figure}[t]
\begin{center}
   \includegraphics[width=1.0\linewidth]{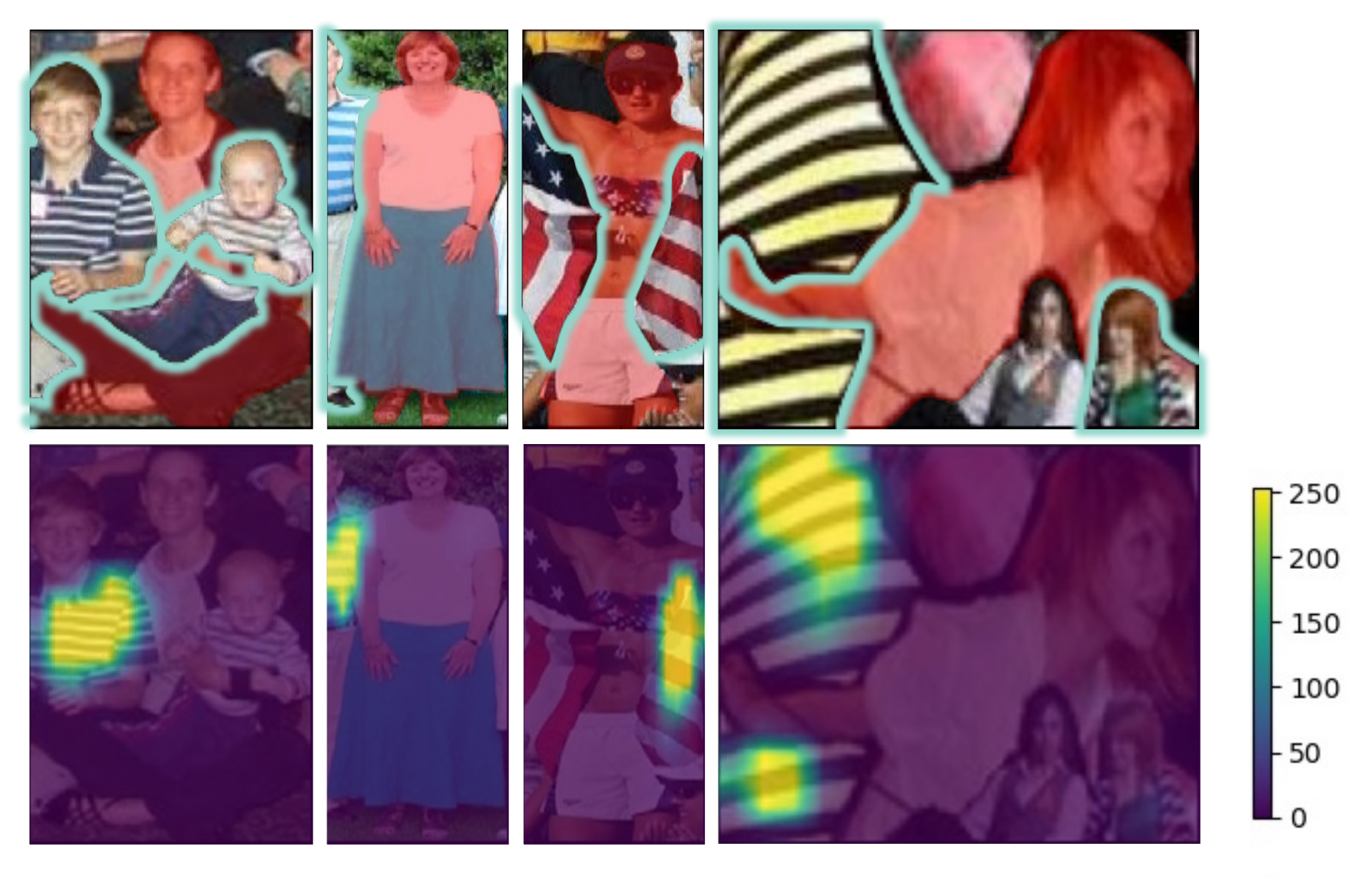}
\end{center}
   \caption{Examples of distractions (best scene in color). \emph{Input images (1st row)} in HAR 
   usually contain target subjects (masked in red) and distractions (enclosed in green contours). We visualize 
   some cases when the attention mechanism mistakenly responds to  \emph{surrounding people (1st and 
   2nd columns)}, \emph{similar objects (3rd column)} and \emph{puzzling backgrounds (4th column)}, 
   leading to \emph{incorrect outputs (2nd row)}.}
\label{fig:fig2}
\end{figure}

To generate qualified features, attribute-related region localization is very crucial, which aims to locate 
the regions that contain useful clues for attribute recognition. With incorrect localization of such regions, 
attribute prediction tends to fail because meaningful features can hardly be captured. In both the surveillance 
and general domains (especially the latter one), the accuracy of attribute localization is usually susceptible 
to pose variations and external occlusions. Therefore, recent attempts develop more sophisticated schemes 
to locate attribute-related regions, including exploiting auxiliary cues of target persons 
\cite{GAM,PGDM,and-or,JLPLS} and applying the attention mechanism \cite{JLPLS,DIAA,HPNet,SRN}. 
They indeed deliver some performance gains; however, it is really a difficult task to obtain accurate human 
auxiliary information, such as detailed body parts and meticulous body poses, in the presence of those 
negative factors. Meanwhile, the traditional attention mechanism is prone to confusing areas, named
distractions, which are caused by surrounding people, similar objects and puzzling backgrounds, as 
shown in Figure \ref{fig:fig2}. These facts suggest space for improvement.

This paper proposes a novel method, namely Distraction-Aware HAR (Da-HAR), dealing with 
prediction of attributes of target persons in the wild (\emph{i.e.} in the general domain). It emphasizes 
the importance of attribute-related region localization and enhances it by a coarse-to-fine attention 
mechanism, which largely reduces irrelevant distraction areas and substantially strengthens the following 
feature learning procedure. Specifically, at the coarse step, a self-mask block is designed to distill
consensus information from extracted CNN features at different scales and highlight the most salient 
regions, in order to learn a rough distraction-aware mask from the input image. At the fine step, the traditional 
attention mechanism \cite{attention} is integrated by introducing a mask branch to further refine the
distraction-aware information. This branch functions in a multi-task manner to boost results of 
classification and segmentation through making use of their interactions. In addition to distraction-aware 
attribute localization, considering that human attributes naturally exist at different semantic levels, we 
jointly use the features extracted from multiple layers of the deep network to improve the discriminative 
power. Comprehensive experiments are carried out on two major public benchmarks, \emph{i.e.} 
WIDER-Attribute and RAP, and state-of-the-art results are reached, which clearly illustrate the 
competency of the proposed approach. 

\section{Related Work}

Due to extensive real-world applications, HAR has drawn many attentions in recent years, with the scores 
on main benchmarks \cite{WIDER,PETA,RAP} continuously improved.

Early work focuses on extracting more discriminative features. \cite{cao2008gender} utilizes Histogram of Oriented 
Gradients (HoG) to represent person appearances and employs an ensemble classifier for gender recognition. 
\cite{joo2013human} builds a rich dictionary of detailed human parts based on HoG and color histograms, handling
more attributes. CNN models, such as GoogLeNet \cite{Inception}, ResNet \cite{ResNet}, and DenseNet \cite{DenseNet}, 
dominate this areas with stronger features and better scores.

Recently, increasing efforts have been made to improve attribute localization and thus boost the HAR performance. 
These methods can be classified into two categories. One explores auxiliary cues of other tasks, \emph{e.g.} body
part detection or annotation \cite{GAM}, pose estimation \cite{and-or,PGDM}, and human parsing \cite{JLPLS},
while the other introduces the attention mechanism to underline more important regions \cite{SRN,HPNet,DIAA,JLPLS}. 

\cite{GAM} leverages body part annotation to calculate the pose-normalized feature maps of the head-shoulder, 
upper body, and lower body parts, respectively. The three feature maps are then employed in prediction and the 
optimal result is combined with that of the entire person to generate final decision. \cite{and-or} jointly infers 
human poses and attributes in a sparse graph, which is built based on the annotated keypoints of the target person. 
\cite{PGDM} transfers the knowledge learned from an off-the-shelf human pose estimation network and integrates 
pose information into attribute recognition to boost the performance. \cite{JLPLS} simultaneously performs 
attribute recognition and human parsing in a multi-task learning manner, which further refines the features by 
parsing results. Unfortunately, target persons in HAR often have arbitrary variations in pose, occlusion, and 
background, \emph{e.g.} in the surveillance and general domains. In this case, the clues of the auxiliary tasks 
mentioned above are no longer accurately available, making those methods unstable.

\begin{figure*}
\begin{center}
  \includegraphics[width=0.9\linewidth]{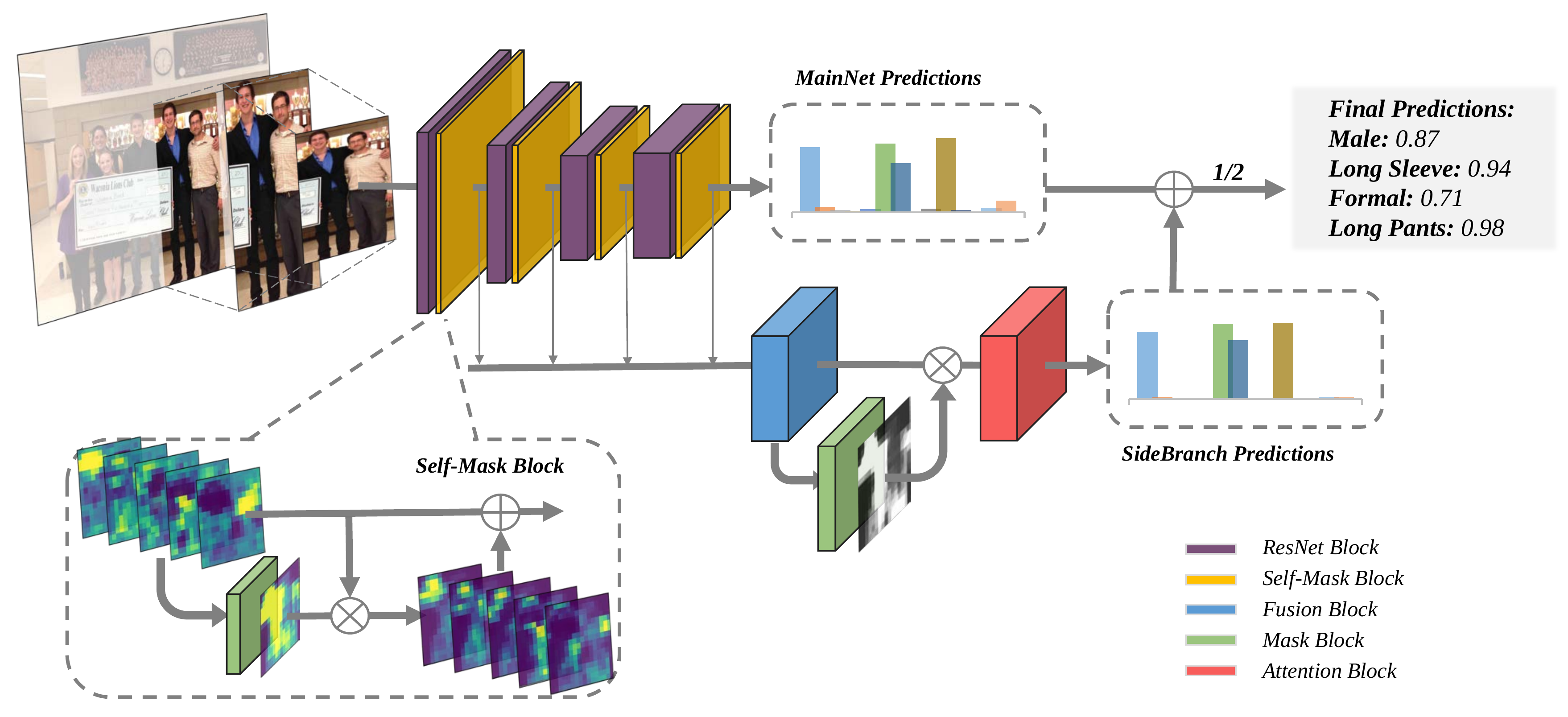}
\end{center}
   \caption{Overall framework of our Da-HAR. The Main Net (\emph{purple}) follows the structure of
   ResNet-101. We insert several \emph{self-mask blocks (yellow)} between different layers in the backbone network 
   to learn the distraction awareness. Features from different levels are collected and fused by a \emph{fusion block (blue)},
   and the result is then sent to a side prediction branch, which contains a \emph{mask block (green)} and an 
   \emph{attention block (red)} to guide the network to better concentrate on the attribute-related regions.}
\label{fig:fig3}
\end{figure*}

\cite{SRN} firstly applies the attention mechanism to attribute recognition and discovers that the confidence-weighted
scheme benefits most to recognition scores. \cite{DIAA} follows this intuition and extends the regular attention 
scheme to a multi-scale attention-aggregated one. \cite{HPNet} learns multiple attentive masks corresponding to 
features at different layers and adds them back on the original features to form up attentive ones. \cite{JLPLS} combines 
parsing attention, label attention, and spatial attention to ameliorate the accuracy of attribute localization. In particular, 
spatial attention learns several attention masks to better concentrate on attribute-related regions for prediction. 
However, these methods conduct attention learning on the entire input patch without any constraint, and the weights 
or masks generated are thus sensitive to confusion areas incurred by surrounding people, similar objects, and puzzling 
background, as shown in Fig. 2, making their results problematic.

The proposed method presents a novel approach, which tends to make use of both the advantages of the two groups of 
methods by inducing additional clues from person segmentation into attention learning in a progressive way. It localizes 
more accurate attribute regions by largely eliminating distraction areas through a coarse-to-fine attention scheme, and 
builds more powerful attribute features by integrating the ones extracted from multiple layers of the deep network. 

\section{Methodology}

\subsection{Overall Framework}

The overall framework of our approach is illustrated in Figure \ref{fig:fig3}. It consists of a main network as well as a 
side branch. Given an image patch with a target person, the proposed method processes it through both the main network 
and the side branch and their attribute predictions are combined to provide final results. 

In the main network, we adopt ResNet-101 as the backbone, following recent studies \cite{SRN,DIAA}. We 
then add the self-mask block to the backbone to generate the features that are enhanced by the learned coarse 
distraction awareness. The features finally pass through a global average pooling layer and deliver Main Net 
predictions. In the side branch, we collect and combine the features at multiple layers from the backbone. 
These aggregated features are fed to the masked attention block, which further refines the previous distraction 
awareness. The same structure for classification is used to produce Side Branch predictions. At last, the 
Main Net and Side Branch predictions are averaged for decision making.

\subsection{Coarse-to-fine Distraction-Aware Learning}

As mentioned in Sec. 1, distractions with the patterns or appearances that are similar to the target may appear in the 
given image in the form of surrounding people, similar objects, and puzzling backgrounds (see Figure \ref{fig:fig2}), 
which tend to induce false activations during the recognition process. The case even degenerates in real-world 
crowded scenes, such as assembly and campaign. Therefore, it is a difficult task to separate the target from distractions 
through bounding boxes. To deal with this issue, we propose a coarse-to-fine attention mechanism to progressively 
learn distraction awareness and substantially strengthen attribute features.

\begin{figure}
\begin{center}
  \includegraphics[width=0.75\linewidth]{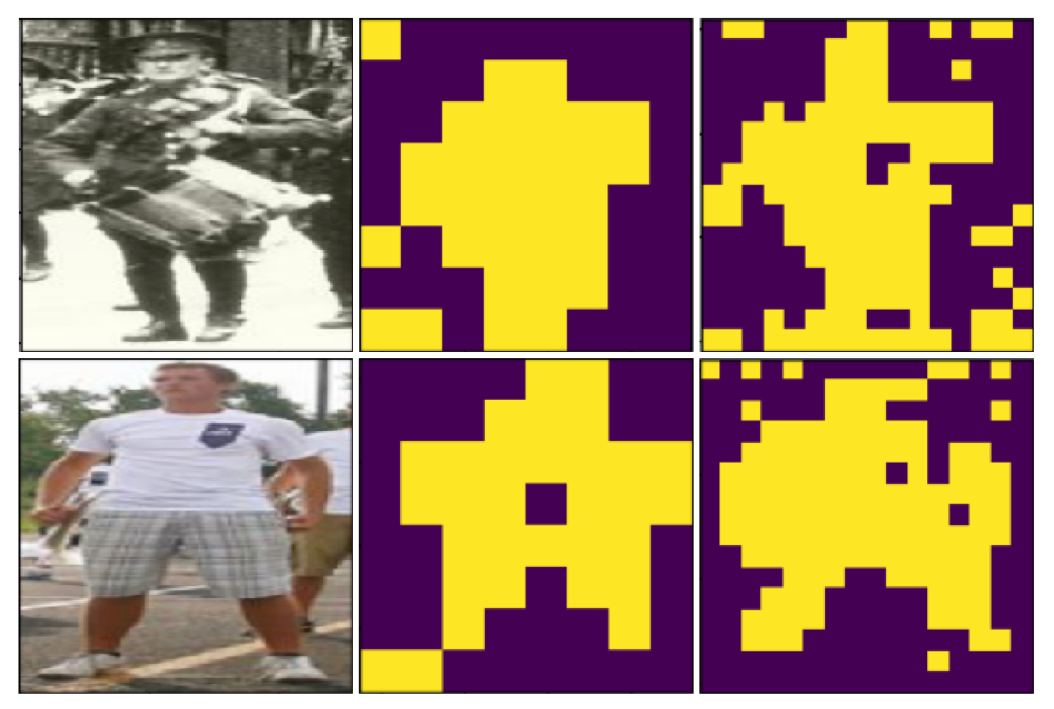}
\end{center}
   \caption{Visualization of summed features. By summing up the output features at the 4th layer (\emph{3rd column}) 
   or the 5th layer (\emph{2nd column}) along the dimension of the channels and using the median value as a threshold 
   for binarization, the approximate outline of the target person is rendered.}
\label{fig:fig4}
\end{figure}

{\bf Coarse Distraction-Aware Learning.} In the first stage, we roughly approximate the location of the target person 
by using the provided image-level labels. As we know, labels are assigned to the target person, to whom the highlighted 
features (activated neurons) are correlated. Therefore, a rough localization can be obtained by combing the features 
according to a weighted sum rule. To verify this intuition, we sum up all the channels of the output feature from the 
4th layer or the 5th layer and employ the median value as a threshold to binarize the summed feature map. As we can 
see in Figure \ref{fig:fig4}, an approximate outline of the target person.

Based on the analysis above, we propose a self-mask block to learn the distraction-awareness. 
The structure of our self-mask block is shown in Figure \ref{fig:fig3}. It contains a mask block to predict the 
saliency information, which is employed to optimize the input features to filter out the unconcerned regions. The 
mask block is constructed similar to a residual block, which is a stack of three $1\times1$ convolution layers 
and some batch normalization and activation layers. Since the self-mask blocks access the features at the 
corresponding layers, they only capture coarse distraction information. 

{\bf Fine Distraction-Aware Learning.} To further improve the localization accuracy of attribute-related regions, 
we propose a masked attention block. Since this block has access to the features collected from different 
layers, which contain the coarse distraction-awareness, it makes refinement to build more precise 
distraction-awareness.

A confidence-weighted attention mechanism \cite{SRN} is exploited in the side branch to guide the network to 
explicitly focus on the attribute-related regions. However, recall that the essence of the attention mechanism is 
a re-weighting process on the give features, and the attention mechanism thus cannot rectify the false 
activations induced by distractions. To introduce distraction-aware learning to the regular attention mechanism, 
we design a masked attention block by integrating a mask branch after multi-level feature fusion (described in 
the next sub-section), as shown in Figure \ref{fig:fig3}. This mask is employed to preprocess the fused feature 
before it is passed to the attention module. Note that the mask block constructed here is different from that in 
the self-mask block. We add two residual blocks on the top of the mask block so that it works in a multi-task 
manner, dealing with recognition and segmentation rather than simply distilling the consensus information.

\subsection{Multi-Level Feature Aggregation}

Features at multiple layers are collected and leveraged in the side branch to enhance the representation 
ability, as human attributes correspond to different semantic levels. To be specific, some attributes, such as long 
hair, wearing logo and stripes, are highly correlated to low-level features, \emph{e.g.} colors, edges and local textons, 
which are encoded by the bottom layers of the network. Meanwhile, other attributes, like wearing T-shirt, 
sunglasses and jeans, are object-related, which are relevant to mid-level semantics extracted from deeper layers. 
In addition, a few attributes are at semantically high level and related to the features from top layers with several 
co-existing evidences. For example, gender is linked to body shapes and dressing styles, while dressing formally 
relates to ties and suits.

As features at different layers possess different spatial sizes and numbers of channels, a fusion block is proposed to 
balance the spatial resolutions and channel widths of these features. This block consists of three components: a 
$1\times1$ convolution layer for channel reduction to the channel widths; a re-sampling layer for scale adjustment to 
reconcile spatial resolutions; and a layer of stacked residual blocks to alleviate the aliasing effect of up-sampling.

\subsection{Training Scheme}

{\bf Loss Functions.} The Binary Cross-Entropy (BCE) loss is adopted in model training, formulated as:
\begin{equation}
\begin{aligned} 
  L_b(\hat{y}_{p},y) = - \sum_{c=1}^{C}log(\sigma(\hat{y}_{p}^{c}))y^{c} + log(1-\sigma(\hat{y}_{p}^{c}))(1-y^{c}),
\end{aligned}
\end{equation}
where $C$ is the total number of attributes, $\hat{y}_p^c$ and $y^c$ correspond to the predicted result and ground-truth 
label for each attribute $c$, respectively, and $\sigma(\cdot)$ represents the sigmoid activation function.

When the total number of concerned attributes increases, the influence of the class imbalance problem can no longer be neglected.
We thus also employ the weighted BCE-loss \cite{WPAL} as:

\begin{equation}
\begin{aligned} 
  L_w(\hat{y}_{p},y) = - \sum_{c=1}^{C}\frac{1}{2\omega_c}log(\sigma(\hat{y}_{p}^{c}))y^{c} + \\
\frac{1}{2(1-\omega_c)} log(1-\sigma(\hat{y}_{p}^{c}))(1-y^{c}),
\end{aligned}
\end{equation}
where $\omega_c$ is the positive sample ratio of attribute $c$.

{\bf Supervision without Ground-Truth.} If the ground-truth masks are given in the HAR datasets, our training can be easily 
performed. Unfortunately, the public HAR datasets rarely provide pixel-level annotations. Besides, pixel-level manual 
annotations are time-consuming. Therefore, we make use of the semantic segmentation results, which are obtained by an 
off-the-shelf segmentation technique, as our pixel-level ground truths, since it reports excellent performance.

\begin{table*}\small
\begin{center}
\resizebox{6.7in}{!}{
\begin{tabular}{lp{14pt}<{\centering}p{14pt}<{\centering}p{14pt}<{\centering}p{14pt}<{\centering}p{14pt}<{\centering}
p{14pt}<{\centering}p{14pt}<{\centering}p{14pt}<{\centering}p{14pt}<{\centering}p{14pt}<{\centering}
p{14pt}<{\centering}p{14pt}<{\centering}p{14pt}<{\centering}p{14pt}<{\centering}|p{36pt}<{\centering}}
\specialrule{0.15em}{3pt}{3pt}
\specialrule{0em}{0pt}{1.5pt}
\normalsize{\bf Method} & 
\rotatebox{65}{Male} & \rotatebox{65}{Long Hair} & \rotatebox{65}{Sunglasses} & \rotatebox{65}{Hat} & 
\rotatebox{65}{T-shirt} & \rotatebox{65}{Long Sleeve} & \rotatebox{65}{Formal} & \rotatebox{65}{Shorts} & 
\rotatebox{65}{Jeans} & \rotatebox{65}{Long Pants} & \rotatebox{65}{Skirts} & \rotatebox{65}{Face Mask} & 
\rotatebox{65}{Logo} & \rotatebox{65}{Plaid} & \normalsize{\bf mAP \tiny{(\%)}} \\
\specialrule{0.1em}{3pt}{2pt}
\footnotesize{\bf Imbalance Ratio} & \footnotesize{1:1} & \footnotesize{1:3} & \footnotesize{1:18} & \footnotesize{1:3} & 
\footnotesize{1:4} & \footnotesize{1:1} & \footnotesize{1:13} & \footnotesize{1:6} & \footnotesize{1:11} & \footnotesize{1:2} & 
\footnotesize{1:9} & \footnotesize{1:28} & \footnotesize{1:3} & \footnotesize{1:18} & \\
\specialrule{0.05em}{1.5pt}{3pt}
R-CNN \tiny\emph{ICCV'15} & 94 & 81 & 60 & 91 & 76 & 94 & 78 & 89 & 68 & 96 & 80 & 72 & 87 & 55 & 80.0 \\
R*CNN \tiny\emph{ICCV'15} & 94 & 82 & 62 & 91 & 76 & 95 & 79 & 89 & 68 & 96 & 80 & 73 & 87 & 56 & 80.5 \\
DHC \tiny\emph{ECCV'16} & 94 & 82 & 64 & 92 & 78 & 95 & 80 & 90 & 69 & 96 & 81 & 76 & 88 & 55 & 81.3 \\
VeSPA \tiny\emph{BMVC'17} & - & - & - & - & - & - & - & - & - & - & - & - & - & - & 82.4 \\
CAM \tiny\emph{PRL'17} & 95 & 85 & 71 & 94 & 78 & 96 & 81 & 89 & 75 & 96 & 81 & 73 & 88 & 60 & 82.9 \\
ResNet-101 \tiny\emph{CVPR'16} & 94 & 85 & 69 & 91 & 80 & 96 & 83 & 91 & 78 & 95 & 82 & 74 & 89 & 65 & 83.7 \\
SRN* \tiny\emph{CVPR'17} & 95 & 87 & 72 & 92 & 82 & 95 & 84 & 92 & 80 & 96 & 84 & 76 & 90 & 66 & 85.1 \\
DIAA \tiny\emph{ECCV'18} & 96 & 88 & 74 & 93 & 83 & 96 & 85 & \bf{93} & \bf{81} & 96 & 85 & 78 & 90 & 68 & 86.4 \\
\specialrule{0.05em}{3pt}{3pt}
Da-HAR \bf{(Ours)} & 
\bf{97} & \bf{89} & \bf{76} & \bf{96} & \bf{85} & \bf{97} & \bf{86} & 
92 & \bf{81} & \bf{97} & \bf{87} & \bf{79} & \bf{91} & \bf{70} & 
\bf{87.3}\\
\specialrule{0.1em}{2pt}{3pt}
\end{tabular}}
\end{center}
\caption{Quantitative comparison in terms of mAP between our proposed approach and eight counterparts on the 
WIDER-Attribute dataset. Note that the asterisk mark next to SRN indicates that the method is re-implemented 
by \cite{DIAA}, because the original work includes the validation set for training while the others do not.}
\label{tab:tab1}
\end{table*}

Specifically, we train the segmentation network on the MS-COCO dataset, where FPN \cite{FPN} is employed 
as the backbone network. The P2 feature is utilized as the final representation for segmentation. We crop target 
persons by bounding boxes and resize them to $224\times224$ as input. For training the masks, we only exploit 
the annotations of the target person and discard all the unnecessary labels. According to our experiments, the 
generated segmentation results well serve as the ground truths for the masked attention block in Da-HAR, except 
for some cases when the target person is mostly occluded by surrounding people or objects.

\section{Experiments}

We evaluate our approach on two major public datasets, namely WIDER-Attribute and RAP. The WIDER-Attribute dataset 
consists of 13,789 images, where 57,524 instances are labeled with corresponding bounding boxes and 14 attribute 
categories. The RAP dataset is smaller than WIDER-Attribute, but it contains a larger number of attribute categories. It has 
a total number of 41,585 cropped images from 26 surveillance cameras, and the images are annotated with 69 attribute 
categories.

\subsection{Training Details} 

We utilize a ResNet-101 model \cite{ResNet} pre-trained on the ImageNet dataset 
\cite{deng2009imagenet}, as the backbone of our Da-HAR. All the proposed blocks, \emph{i.e.}, the self-mask 
block, the feature fusion block, and the masked attention block, are initialized with Gaussian noise ($m=0, \sigma=0.01$). 

To avoid over-fitting, the data augmentation strategies employed in \cite{SRN} are adopted in our training. We 
firstly resize the input images to $256\times256$. Then, the resized images are cropped at four corners and the 
center, with the cropping width and height randomly chosen from the set \{256, 224, 192, 168, 128\}. At last, 
the cropped images are resized to $224\times224$. Note that we keep the ratio of height/width at 2 for the 
RAP dataset. Random horizontal flipping and color jittering are also applied for the data augmentation.

The stochastic gradient descent algorithm is utilized in the training process, with a batch size of 32, a momentum 
of 0.9 and a weight decay of 0.0005. The initial learning rate is set to 0.003, and gamma is set to 0.1. Our model is 
trained on a single NVIDIA 1080Ti GPU. Two different strategies are used in inference. One directly resizes the 
input images to $256\times256$, while the other introduces an averaged result of five-crop evaluation.

\subsection{Results}

{\bf WIDER-Attribute.} On this dataset, mean Average Precision (mAP) is employed as the metric. We compare our 
Da-HAR with the latest methods, including R-CNN \cite{R-CNN}, R*CNN \cite{R*CNN}, DHC \cite{WIDER}, 
CAM \cite{CAM}, VeSPA \cite{VeSPA}, SRN \cite{SRN}, DIAA \cite{DIAA} and a fine-tuned ResNet-101 
network \cite{ResNet}. Note that SRN results are quoted from \cite{DIAA}, where the method is re-implemented 
without using the validation set for fair comparison.

Table \ref{tab:tab1} displays the results of different methods in details. We can see that our approach outperforms 
all the existing methods on the WIDER-Attribute dataset. Compared to DIAA, which reports the previous 
state-of-the-art score, Da-HAR achieves an mAP gain of 0.9\%. Specifically, the results of some attributes related 
to accessories (\emph{e.g.}, Hat and Sunglasses) and with strong local patterns (\emph{e.g.}, T-shirt, Skirt and 
Plaid) increase more than 1\%. Since these attributes usually have strong activations within small regions, they are 
easier to be influenced by surrounding distractions. By introducing the coarse-to-fine attention mechanism, 
Da-HAR proves effective in reducing such interferences and thus strengthening features. In particular, the proposed 
method shows the largest improvement (3\%) on the Hat attribute, even though the baseline is relatively high 
(93\%). In general, Hat is a frequently appeared attribute and its prediction seriously suffers from distractions. 
It highlights the advantage of the proposed method.

\begin{table}\small
\begin{center}
\begin{tabular}{p{55pt}|p{11.5pt}<{\centering}|p{16.5pt}<{\centering}p{16.5pt}<{\centering}p{16.5pt}<{\centering}p{16.5pt}<{\centering}p{16.5pt}<{\centering}}
\specialrule{0.15em}{3pt}{3pt}
\specialrule{0em}{0pt}{1.5pt}
{\normalsize{\bf Method}} & \footnotesize{\emph{loss}} & \normalsize{\emph{mA \tiny{(\%)}}} & 
\normalsize{\emph{Acc \tiny{(\%)}}}  & \normalsize{\emph{Prec \tiny{(\%)}}} & 
\normalsize{\emph{Rec \tiny{(\%)}}} & \normalsize{\emph{F1 \tiny{(\%)}}} \\
\specialrule{0.1em}{3pt}{3pt}
HPNet \tiny\emph{ICCV'17} & \footnotesize{\emph{s.}} & 76.12 & 65.39 & 77.53 & 78.79 & 78.05\\
JRL \tiny\emph{ICCV'17} & \footnotesize{\emph{o.}} & 77.81 & - & 78.11 & 78.98 & 78.58\\
VeSPA \tiny\emph{BMVC'17} & \footnotesize{\emph{s.}} & 77.70 & 67.35 & 79.51 & 79.67 & 79.59\\
WPAL \tiny\emph{BMVC'17} & \footnotesize{\emph{w.}} & 81.25 & 50.30 & 57.17 & 78.39 & 66.12\\
GAM \tiny\emph{AVSS'17} & \footnotesize{\emph{o.}} & 79.73 & \bf{83.79} & 76.96 & 78.72 & 77.83 \\
GRL \tiny\emph{IJCAI'18} & \footnotesize{\emph{w.}} & 81.20 & - & 77.70 & 80.90 & 79.29\\
LGNet \tiny\emph{BMVC'18} & \footnotesize{\emph{s.}} & 78.68 & 68.00 & 80.36 & 79.82 & 80.09\\
PGDM \tiny\emph{ICME'18} & \footnotesize{\emph{w.}} & 74.31 & 64.57 & 78.86 & 75.90 & 77.35\\
VSGR \tiny\emph{AAAI'19} & \footnotesize{\emph{o.}} & 77.91 & \underline{70.04} & 82.05 & 80.64 & \bf{81.34} \\
RCRA \tiny\emph{AAAI'19} & \footnotesize{\emph{w.}} & 78.47 & - & \underline{82.67} & 76.65 & 79.54 \\
IA$^2$Net \tiny\emph{PRL'19} & \footnotesize{\emph{f.}} & 77.44 & 67.75 & 79.01 & 77.45 & 78.03 \\
JLPLS \tiny\emph{TIP'19} & \footnotesize{\emph{o.}} & 81.25 & 67.91 & 78.56 & \underline{81.45} & 79.98 \\
CoCNN \tiny\emph{IJCAI'19} & \footnotesize{\emph{o.}} & 81.42 & 68.37 & 81.04 & 80.27 & 80.65 \\
DCL \tiny\emph{ICCV'19} & \footnotesize{\emph{m.}} & \underline{83.70} & - & - & - & - \\
\specialrule{0.05em}{3pt}{3pt}
\multirow{3}{*}{Da-HAR {\bf (Ours)}} & \footnotesize{\emph{o.}} & 73.78 & 68.67 & \bf{84.54} & 76.84 & 80.50 \\
& \footnotesize{\emph{w.}} & \bf{84.28} & 59.84 & 66.50 & \bf{84.13} & 74.28 \\
& \footnotesize{\emph{m.}} & 79.44 & 68.86 & 80.14 & 81.30 & \underline{80.72} \\
\specialrule{0.1em}{2pt}{3pt}
\end{tabular}
\end{center}
\caption{Quantitative results of our proposed method and fourteen counterparts on the RAP dataset. \emph{s.}, 
\emph{o.}, \emph{w.}, \emph{f.}, and \emph{m.} represent the Softmax loss, Original BCE-loss, Weighted 
BCE-loss, Focal BCE-loss, and a Mixed loss, respectively.} 
\label{tab:tab3}
\end{table}

{\bf RAP.} On this dataset, we follow the standard protocol \cite{RAP} and only report the prediction results on 51 
attributes whose positive sample ratios are higher than 1\%. Multiple metrics are used, involving mean Accuracy (mA), 
instance Accuracy (Acc), instance Precision (Prec), instance Recall (Rec), and instance F1 score (F1). Refer to \cite{RAP}
for their definitions and explanations. We compare our approach with 14 existing counterparts, including HPNet 
\cite{HPNet}, JRL \cite{JRL}, VeSPA \cite{VeSPA}, WPAL \cite{WPAL}, GAM \cite{GAM}, GRL \cite{GRL}, 
LGNet \cite{LGNet}, PGDM \cite{PGDM}, VSGR \cite{VSGR}, RCRA \cite{RCRA}, I$^2$ANet \cite{IA2Net}, 
JLPLS \cite{JLPLS}, CoCNN \cite{CoCNN}, and DCL \cite{DCL}, as shown in Table \ref{tab:tab3}. The samples in 
the RAP dataset are collected from real world surveillance scenarios, and compared to the ones in WIDER-Attribute, 
there are less distractions. Under this circumstance, our method still reaches very promising results compared to the 
state-of-the-art methods.

Since the problem of class imbalance in RAP is very severe, the performance fluctuates under different loss 
functions. Da-HAR effectively filters out distractions and thus reduces false positives, showing the 
superiority in Precision with the highest score of 84.54\% under BCE-loss. As the original BCE-loss 
emphasizes the majority attributes, the trained network tends to induce strong biases on the minority 
ones, which impairs Recall. The weighted BCE-loss assigns penalties to the minority attributes and enforces 
the network to better distinguish them. In this case, the highest instance Recall of 84.13\% is achieved 
, along with the highest mean Accuracy of 84.28\%. However, the discriminative information 
in the majority attributes is suppressed, which leads to a Precision drop. At last, a mixed loss function 
makes a better trade-off between Precision and Recall and delivers a moderate F1 score of 80.72\%. 

\subsection{Ablation Study}

\begin{table}\small
\begin{center}
\resizebox{3.22in}{!}{
\begin{tabular}{cccc|c}
\specialrule{0.15em}{3pt}{3pt}
{\footnotesize Multi-Level} & \multirow{2}{*}{\footnotesize Self-Mask} & 
{\footnotesize Masked} & {\footnotesize Ignore} & \bf{mAP \tiny{(\%)}} \\
{\footnotesize Feature} & & {\footnotesize Attention} & {\footnotesize 0-labeled} & {\scriptsize Crops / Full} \\
\specialrule{0.1em}{3pt}{3pt}
& & & & 85.3 / 86.2 \\
\normalsize{\checkmark} & & & & 85.7 / 86.5 \\
& \normalsize{\checkmark} & & & 85.8 / 86.6 \\
& & \normalsize{\checkmark} & & 86.0 / 86.8 \\
\normalsize{\checkmark} & \normalsize{\checkmark} & \normalsize{\checkmark} & & 86.2 / 87.1 \\
\normalsize{\checkmark} & \normalsize{\checkmark} & \normalsize{\checkmark} & \normalsize{\checkmark} & 86.5 / \bf{87.3} \\
\specialrule{0.05em}{3pt}{3pt}
\normalsize{\checkmark} & \normalsize{\checkmark} & \normalsize{\checkmark} & \normalsize{\checkmark} & \bf{87.2 / 88.0} \\
\specialrule{0.1em}{2pt}{3pt}
\end{tabular}}
\end{center}
\caption{Ablation Study on the WIDER-Attribute dataset. We re-implement SRN \cite{SRN} without its spatial 
regularization module as our baseline, whose results are reported in the first row. The last row displays the results 
obtained by adding the validation subset to the training process.}
\label{tab:tab2}
\end{table}

To validate our contributions, we further perform ablation studies on the WIDER-Attribute dataset. The images are 
acquired in unconstrained scenarios, including marching, parading, cheering and different ceremonies, where the 
target person is more likely to appear in crowds, \emph{i.e.}, distractions occur more frequently in cropped input images. 
Therefore, this dataset is suitable for analyzing the proposed method.

{\bf Sub-Module Analysis.} Here, we investigate the contribution of each module to the final results of Da-HAR 
in Table \ref{tab:tab2}. We re-implement SRN \cite{SRN} without its spatial regularization module to serve as our 
baseline network, which achieves an mAP of 85.3\% (obtained by using the average results of five-crop evaluation).
If we utilize the uncropped (full) patch and resize it to $256\times256$ as the input, the baseline increases to 86.2\%, 
as the uncropped patch contains more information. Our feature fusion block boosts the performance to 85.7\%/86.5\%, 
which demonstrates the effectiveness of leveraging multi-level semantic clues. When the self-mask block is applied 
only, an mAP of 85.8\%/86.6\% is reached by learning the coarse distraction-awareness. The masked attention block, 
\emph{i.e.}, the fine distraction-aware learning step, further improves the mAP to 86.0\%/86.8\%. These facts 
underline the necessity of the coarse-to-fine attention scheme. By integrating all the three components, the mAP rises 
to 86.2\%/87.1\%. We also observe that the mAP score grows to 86.5\%/87.3\%, if we ignore 0-label attributes at 
the training phase, which are previously treated as negatives in the literature. This step makes data more consistent 
in training, in spite of losing some samples. If the validation set is added to the training one as \cite{SRN} does, an 
mAP of 87.2\%/88.0\% is obtained. 

\begin{figure*}
\begin{center}
  \includegraphics[height=245pt]{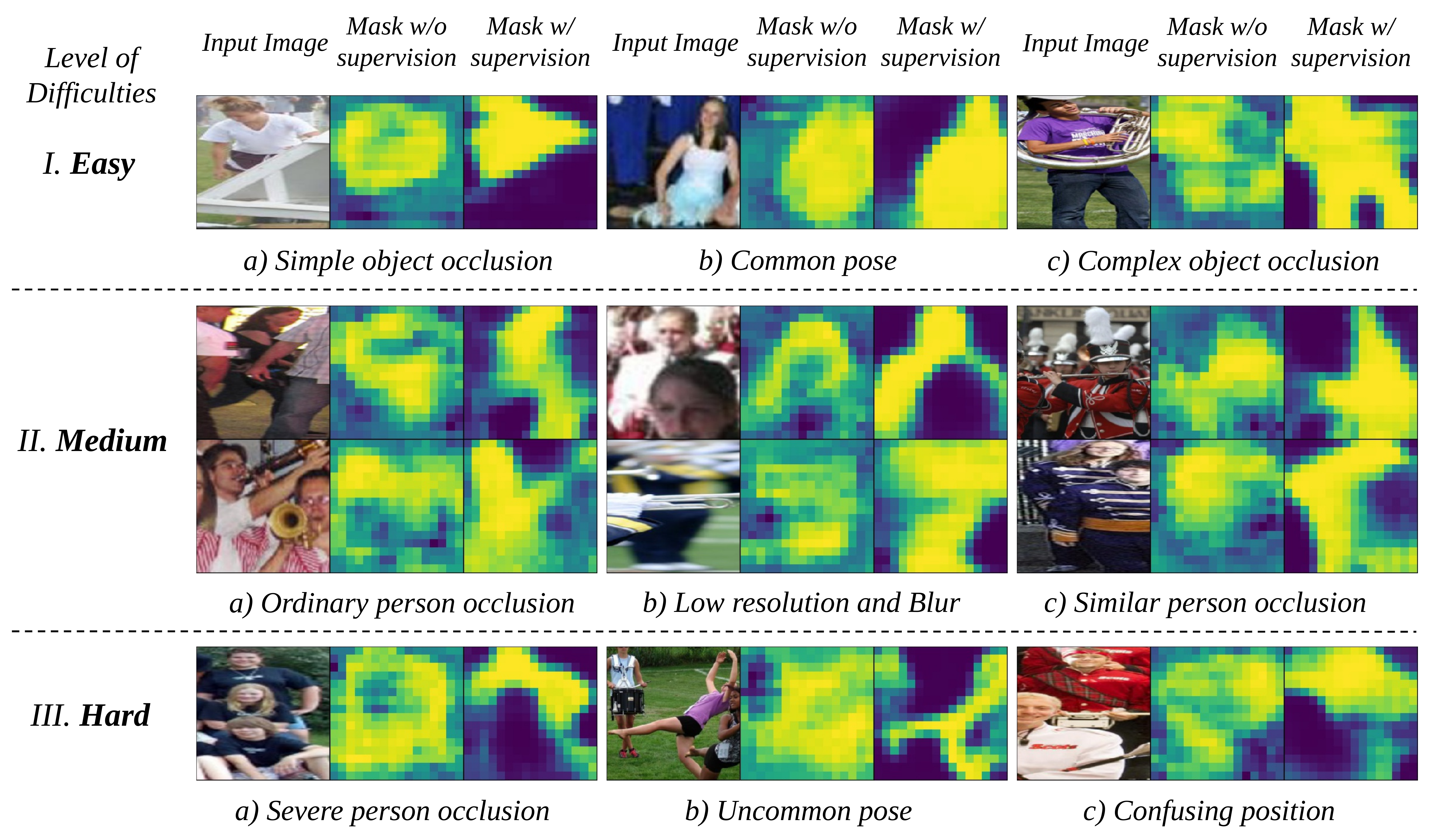}
\end{center}
   \caption{Visualization of Attentive Mask. The visualized masks are trained with and without the auxiliary supervision.}
\label{fig:fig5}
\end{figure*}

{\bf Masked Attention Analysis.} To further verify the effectiveness of the coarse-to-fine distraction-aware feature 
learning and demonstrate the impact of the auxiliary supervision from segmentation, we visualize the saliency mask, 
\emph{i.e.}, the output of the masked attention module in the side branch in Figure \ref{fig:fig5}. The visualized 
masks are trained with and without the auxiliary supervision. For convenience, a cropped input image and the 
corresponding masks trained without/with supervision are gathered to form a triplet of images. These triplets are 
arranged into three panels, which are separated by two dashed lines, according to their different levels of difficulties. 
In each panel, they are grouped into three zones, with respect to the challenges. 

The first panel consists of images with simple object occlusion (\emph{I-a}), common pose (\emph{I-b}), and 
moderate occlusion by complex objects (\emph{I-c}). In this case, the saliency masks generated without 
supervision are well estimated with slight noises. When we use proper supervision, the generated masks 
are obviously better.

The second panel contains images with severe occlusions by ordinary people (\emph{II-a}), low resolution and 
blur (\emph{II-b}), and occlusions by people having similar appearances (\emph{II-c}). Although the distractions 
are more severe compared to those in the first panel, our mask still locates the target person without supervision. 
If supervision is included in training, our method achieves a very accurate localization.

The images in the third panel are considered to be hard for HAR with heavy occlusions by people having extremely 
similar appearances (\emph{III-a}), irregular pose (\emph{III-b}), and severe occlusions with confusing positioning 
(\emph{III-c}). The white jacket (\emph{III-c}) are wrongly identified as the pants of the target person, who is 
dressed in red. Under such circumstances, the masks trained without supervision are not good, but the supervised 
ones are decent, which demonstrates the effectiveness of the pre-trained segmentation network.

{\bf Computational Complexity Analysis}. To better understand our method, we calculate the overall network parameters 
and record the inference time. The baseline network has 45.4M parameters, and the inference time of a single image by 
a NVIDIA 1080Ti is 22.28ms. When all the blocks are added, the final network (Da-HAR) has 52.4M parameters in 
total and the inference time only slightly increases to 24.04ms.

\section{Conclusion}

In this paper, we introduce a distraction-aware learning method for HAR, namely Da-HAR. It underlines the necessity 
of attribute-related region localization and a coarse-to-fine attention scheme is proposed for this issue. A self-mask
block is presented to roughly locate distractions at each layer, and a masked attention mechanism is designed to refine 
the coarse distraction awareness to further filter out false activations. We also integrate the multi-level semantics of 
human attributes to improve the performance. Extensive experiments are carried out on the WIDER-Attribute and RAP 
datasets and state of the art results are reached, which demonstrate the effectiveness of the proposed Da-HAR.

\section{Acknowledgment}

This work was partly supported by the National Key Research and Development Plan (No. 2016YFB1001002), the 
National Natural Science Foundation of China (No. 61802391), and the Fundamental Research Funds for the Central 
Universities.


{\small
\bibliography{egbib}
\bibliographystyle{aaai}
}

\end{document}